\documentclass[cameraready]{Interspeech}
\title{Evaluating and Preserving Lexical Stress in English-to-Chinese Speech-to-Speech Translation}

\author[affiliation={1,2}]{Yuchen}{Song}
\author[affiliation={1}]{Xi}{Chen}
\author[affiliation={1}]{Mingze}{Li}
\author[affiliation={1}, correspondingauthor]{Satoshi}{Nakamura}

\address{
    $^1$ The Chinese University of Hong Kong, Shenzhen, China \\
    $^2$ Shenzhen Loop Area Institute, China
}

\email{yuchensong1@link.cuhk.edu.cn, xichen7@link.cuhk.edu.cn, mingzeli1@link.cuhk.edu.cn, snakamura@cuhk.edu.cn}


\usepackage{comment}
\usepackage{float}

\begin{document}

\maketitle

\begin{abstract}
    Speech-to-speech translation (S2ST) systems have achieved impressive progress in semantic accuracy and speech naturalness. However, the cross-lingual transfer of lexical stress—a vital cue for emphasis and speaker intent—remains heavily underexplored, compounded by a lack of reliable automatic evaluation metrics for tonal languages like Chinese. We investigate English-to-Chinese S2ST stress transfer by constructing a stress-annotated Chinese dataset and an XLS-R-based Mandarin stress detector. Integrating this with the English EmphAssess system, we propose a novel objective metric for cross-lingual stress evaluation. Furthermore, we fine-tune CosyVoice3 to build a stress-aware S2ST system. Experiments demonstrate that our proposed S2ST architecture significantly outperforms existing systems in stress translation capability while maintaining competitive translation quality. Furthermore, our evaluation metric exhibits a strong correlation with human subjective judgments.
\end{abstract}

\section{Introduction}

Speech-to-Speech Translation (S2ST) has advanced rapidly in recent years, driven by progress in neural speech modeling and large-scale multilingual systems \cite{sarim2025direct, gupta2025direct, jia2019direct, fung2004grammar}. Recent unified and end-to-end frameworks, such as SeamlessM4T \cite{barrault2023seamlessm4t, dong2023polyvoice,chen2023speech} and UnitY \cite{inaguma2023unity}, have significantly improved semantic accuracy and speech naturalness. Evaluation metrics tailored for S2ST, such as BLASER \cite{chen2023blaser}, further enable robust text-free assessment of translation quality. Modern systems are therefore highly capable of achieving strong semantic fidelity and high-quality audio synthesis. However, an important aspect of natural spoken communication remains insufficiently explored: lexical stress.

Prosody, particularly lexical and phrasal stress, plays a crucial role in conveying emphasis, speaker intent, and pragmatic meaning \cite{dahan2015prosody, werner1994prosodic, tsiamas2024speech, yu2024effects}. Recent work has shown that current speech translation systems only weakly leverage prosodic information during translation \cite{tsiamas2024speech}. In cross-lingual scenarios, incorrect or missing stress may lead to speech that is semantically correct but pragmatically misleading. Despite its importance, stress transfer remains underexplored in S2ST research \cite{do2018sequence, vazquez2025hubert}.

Several recent studies have begun investigating emphasis preservation in speech translation. EmphAssess \cite{de2024emphassess} introduced a benchmark for evaluating emphasis transfer and demonstrated that even state-of-the-art models struggle to reliably preserve stress patterns. StressTransfer \cite{chen2025stresstransfer} further explored stress-aware S2ST architectures, making early progress in this direction. However, evaluating and transferring stress—particularly for tonal languages like Mandarin Chinese—remains a fundamental bottleneck. In Mandarin, stress interacts closely with lexical tone and contextual prosody, resulting in perceptual and acoustic mechanisms that differ drastically from those of stress-accent languages like English. Consequently, the direct migration of English-centric emphasis detection models is infeasible \cite{yu2024effects, he2022automatic, zou2025machine}. Furthermore, even recent advancements in stress transfer for Indic languages, such as Hindi \cite{narasinga2024attempt, akarsh2025examining}, cannot be readily applied due to Mandarin's unique prosodic characteristics. Consequently, accurately evaluating and quantifying stress transfer in English-to-Chinese S2ST remains a critical, unresolved challenge.

On the synthesis side, generating high-quality emphasized speech in the target language is a prerequisite for effective S2ST. Previous works, such as StressTest \cite{yosha2025stresstest}, have successfully utilized models like OpenAI TTS to generate English emphasized speech. However, generating Mandarin stressed speech remains a significant challenge. While recent expressive Audio-LLMs like CosyVoice \cite{du2024cosyvoice} support diverse speech generation, their ability to consistently and reliably render specific lexical stress in Chinese is highly unstable. We identify that the primary obstacle to resolving this bottleneck in English-to-Chinese stress translation is the absence of a high-quality, stress-annotated Chinese speech dataset. Without targeted emphasis data, S2ST pipelines relying on general-purpose TTS models suffer from weakened realization of emphasis.

To address these challenges, we focus on stress modeling and evaluation in English-to-Chinese S2ST. By leveraging self-supervised XLS-R representations \cite{babu2021xls}, we develop a robust framework for stress evaluation and generation. 

The primary contributions of this paper are threefold:
\begin{itemize}
    \item We construct a high-quality, stress-annotated Chinese speech dataset, addressing the critical data scarcity in Mandarin emphasis modeling.
    \item We develop a Mandarin stress detection model named Syl-BiLSTM. Experiments demonstrate that it significantly outperforms English-centric baselines in stress detection tasks. Based on this, we propose an objective metric for cross-lingual stress transfer evaluation, which exhibits a strong correlation with human subjective judgments.
    \item Utilizing our dataset, we develop a stress-aware TTS system and propose an English-to-Chinese stress-aware S2ST architecture. Experimental results indicate that our system significantly surpasses baseline models in stress preservation capability while maintaining competitive translation quality.
\end{itemize}

% \begin{figure*}[t]
%     \centering
%     \includegraphics[width=\linewidth]{architecture.png} % 请确保文件名与上传一致
%     \caption{The architecture of our proposed emphasis-aware speech-to-speech translation system.}
%     \label{fig:model_arch}
% \end{figure*}
% 单栏插入图片，对应附图1

\section{Dataset Construction}

Due to the lack of high-quality Chinese datasets specifically designed for emphasis modeling, we constructed a dedicated dataset following the methodology proposed by Chen et al. \cite{chen2025stresstransfer}. Our data sourcing strategy leverages two subsets: the \textbf{EmphST-Bench} and the \textbf{EmphST-INSTRUCT} corpora. Specifically, we utilized all 218 Chinese samples from EmphST-Bench and selected 200 samples from EmphST-INSTRUCT.

\textbf{Recording protocol.} We recruited two native Mandarin speakers with standard Mandarin proficiency. Each speaker recorded the selected Chinese sentences in a quiet room using a microphone at a 16,000 Hz sampling rate. For each sentence, speakers were shown the target emphasis position and were instructed to produce natural perceived prominence on the marked syllable/character while preserving the lexical tone category and avoiding exaggerated pitch changes or segmental modifications. Each sentence was recorded 3--5 times with the same target emphasis position, varying only the intended stress strength under weak, medium, and strong stress prompts.

\textbf{Quality control.} We manually inspected all recordings and retained only utterances satisfying the following criteria: (i) the intended stressed syllable/character was clearly perceived, (ii) no unintended neighboring stress dominated the target stress, (iii) lexical tones and segmental content were correctly produced, (iv) the audio contained no clipping, noise, or truncation, and (v) forced alignment succeeded at the character level. Recordings failing any criterion were discarded. This process resulted in 1,883 samples covering 418 unique sentences and 2.74 hours of speech. The statistics of our collected dataset are summarized in Table \ref{tab:dataset_stats}.

\begin{table}[H]
    \centering
    \caption{Statistics of the Collected Chinese Stress Dataset.}
    \label{tab:dataset_stats}
    \begin{tabular}{l|c}
        \toprule
        \textbf{Statistic} & \textbf{Value} \\
        \midrule
        Number of Unique Sentences & 418 \\
        Number of Samples & 1883 \\
        Total Hours & 2.74 \\
        Language & Chinese \\
        Speakers & 2 \\
        Avg. Source Text Length (words) & 9.10 \\
        Avg. Target Text Length (chars) & 13.49 \\
        \bottomrule
    \end{tabular}
\end{table}

\section{Cross-Lingual Stress Evaluation Framework}
\label{sec:stress_pipeline}

In this section, we introduce our framework for detecting and evaluating cross-lingual stress transfer. A core challenge in evaluating stress preservation is the fundamental difference between stress-accent languages such as English and tonal languages such as Mandarin Chinese. To address this, we designed a dedicated Mandarin stress detection model, Syl-BiLSTM, and integrated it into a comprehensive cross-lingual evaluation pipeline.

\subsection{Mandarin Stress Detection: Syl-BiLSTM}

\begin{figure}[t]
    \centering
    % 请确保图片文件 train.png 和你的 .tex 文件在同一目录
    \includegraphics[width=0.75\linewidth]{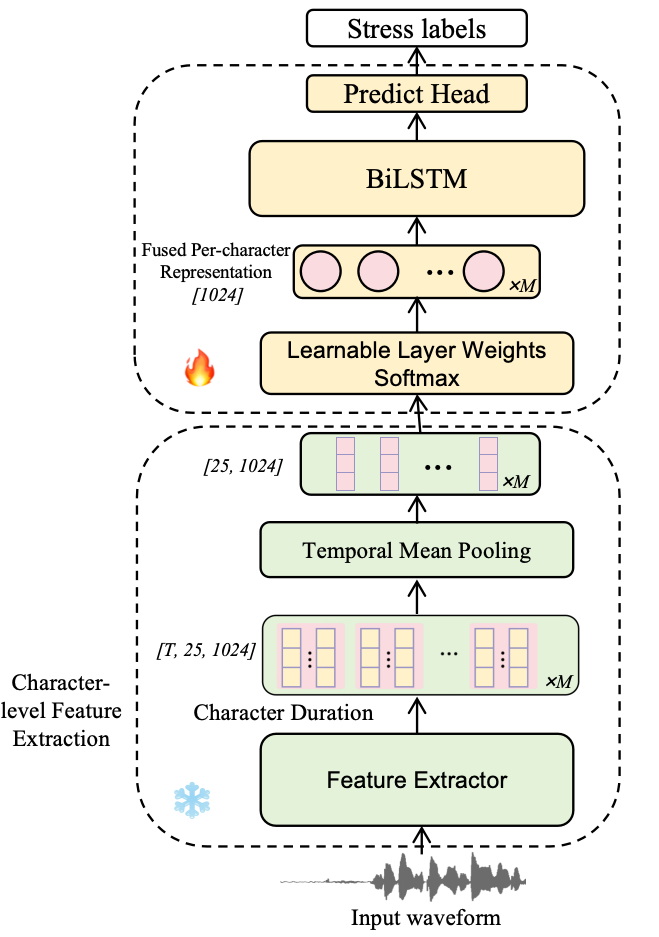}
    \vspace{-0.2cm} % 向上压缩图片与Caption的间距
    \caption{Architecture of the proposed Syl-BiLSTM for Mandarin stress detection.}
    \label{fig:syl_bilstm}
    \vspace{-0.4cm} % 向下压缩Caption与正文的间距
\end{figure}

To accurately detect stress in Mandarin Chinese, we propose Syl-BiLSTM, an architecture operating strictly at the syllable level while leveraging bidirectional contextual modeling (illustrated in Figure \ref{fig:syl_bilstm}).

We extract frame-level representations from all 25 layers of a pre-trained XLS-R model \cite{babu2021xls}, since different layers encode complementary acoustic, prosodic, and semantic cues \cite{pasad2021layer}. Let $M$ denote the number of characters. Using forced-alignment timestamps, we segment the hidden states into $M$ character-specific tensors of size $[T, 25, 1024]$, where $T$ is the variable number of frames per character. We then apply temporal mean pooling over $T$ to obtain a fixed-size syllable-level representation of size $[25, 1024]$ for each character.

The character representations are stacked into an utterance-level sequence of length $M$. A learnable softmax-normalized layer fusion combines the 25 XLS-R layers into a single sequence representation, $H_{\text{fused}} = \sum_{i=1}^{25} \text{softmax}(w_i) \cdot h_i$, of size $[M, 1024]$. To model contextual stress patterns in Mandarin, the fused sequence is passed through a Bidirectional LSTM (BiLSTM), followed by a linear head that predicts a binary stress label ($0$ for unstressed, $1$ for stressed) for each character.

\begin{figure*}[t] 
    \centering
    \includegraphics[width=0.92\textwidth]{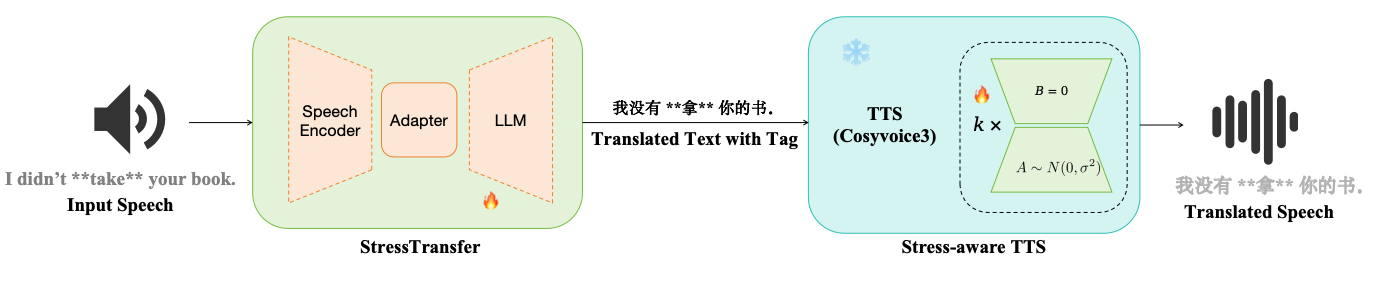} 
    \vspace{-0.2cm} % 压缩间距
    \caption{The overall architecture of our proposed stress-aware S2ST system.}
    \label{fig:model_arch}
    \vspace{-0.4cm} % 压缩间距
\end{figure*}

\subsection{Cross-Lingual Stress Evaluation Pipeline}

Building upon our detection models, we designed a comprehensive cross-lingual evaluation pipeline, which we name the Cross-lingual Emphasis Transfer Score (CETS), to assess the accuracy of emphasis transfer. This evaluation process systematically maps source-side English stress to target-side Chinese translations. 

First, we employ the EmphaClass model \cite{de2024emphassess} to detect emphasis in the source English audio. EmphaClass utilizes XLS-R to generate frame-level binary classifications, and a source word is classified as stressed if more than 50\% of its frames are predicted as emphasized. For the target Chinese speech, we first utilize the Whisper Large model \cite{radford2023robust} to transcribe the synthesized audio into text. Subsequently, we employ the \texttt{fa-zh} forced alignment tool to extract accurate character-level timestamps. With the transcription and boundaries established, our trained Syl-BiLSTM processes the Chinese speech to generate a corresponding binary stress prediction for each character.

To establish an accurate correspondence between the source and target emphasis, we utilize the SimAlign framework \cite{sabet2020simalign} for cross-lingual word alignment. SimAlign leverages multilingual pre-trained language models to automatically extract optimal matching between English source words and Chinese target characters. Based on this precise cross-lingual mapping, we define a strict criterion for successful stress transfer: a stress translation is deemed successful if and only if an English source word detected as stressed by EmphaClass is directly aligned with a target Chinese character detected as stressed by our Syl-BiLSTM. This rigorous matching requirement ensures that the emphasis is not merely generated arbitrarily in the output, but is accurately preserved and transferred to the semantically corresponding position in the translated speech.

\section{Stress-Aware S2ST Architecture}
\label{sec:architecture}

To achieve high-fidelity stress preservation during cross-lingual transfer, we propose a cascaded emphasis-aware speech-to-speech translation (S2ST) framework. As illustrated in Figure \ref{fig:model_arch}, our system bridges an emphasis-aware Speech-to-Text (S2T) translation module with a stress-controllable Text-to-Speech (TTS) synthesis module.

\subsection{Emphasis-Aware Speech Translation}
For the translation component, we adopt the StressTransfer S2TT backbone \cite{chen2025stresstransfer}, using Whisper-Large-v3 as the speech encoder and Qwen2.5-3B as the LLM. The LLM is fine-tuned to perform simultaneous translation and emphasis prediction, outputting the translated Chinese text with explicit stress tags. This explicit tagging bridges the gap between source acoustic emphasis and target linguistic semantics. To maintain general translation capabilities while adapting to the stress prediction task, we apply Low-Rank Adaptation (LoRA) \cite{hu2021lora} to both the speech encoder and the LLM backbone, while fully fine-tuning the adapter module.

\subsection{Stress-Controllable Speech Synthesis}
The synthesized speech is generated by our synthesis module based on the tagged text. We employ CosyVoice 3 \cite{du2024cosyvoice} as our base TTS model. While general-purpose TTS models generate high-quality speech, they often struggle to reliably render specific lexical stress in Mandarin. To address this, we fine-tune CosyVoice 3 on our newly collected Chinese stress-annotated dataset. 

To mitigate catastrophic forgetting and overfitting given the limited size of our dataset, we freeze the text and speech tokenizers and apply LoRA to the attention and projection layers (\texttt{q\_proj}, \texttt{k\_proj}, \texttt{v\_proj}, \texttt{o\_proj}) of the transformer blocks. During training, we explicitly utilize the \texttt{<strong>} tag in the text input to denote stress. The model is optimized using the cross-entropy loss to maximize the probability of predicting the target speech tokens $s_t$, conditioned on the prompt $T_{prompt}$, text $T_{text}$, and the stress tag $P_{tag}$:
\begin{equation}
    \mathcal{L} = -\sum_{t} \log P(s_t \mid s_{<t}, T_{prompt}, T_{text}, P_{tag})
    \label{eq:loss}
\end{equation}
This targeted fine-tuning enables the TTS model to robustly interpret the stress tags and generate natural Mandarin speech with accurate emphasis.

\section{Experiments}
\label{sec:experiments}

\subsection{Experimental Setup}
To systematically evaluate our proposed framework, we utilize our collected Chinese stress-annotated dataset. For each evaluation run, we randomly sample 35 unique sentences from the EmphST-Bench subset to construct the test set, yielding approximately 170 emphasized speech samples ($\sim$10\% of the entire dataset). To ensure statistical robustness, this random sampling process is repeated 5 times. All reported objective results are averaged over these 5 random sentence splits, presented as mean $\pm$ standard deviation. (Details regarding the data sampling for the subjective evaluation are provided in Section \ref{subsec:subjective}).

The two-speaker Mandarin stress corpus is not used to train the StressTransfer-based S2TT module. During S2ST inference, the TTS module synthesizes from stress-tagged Mandarin text using a fixed default voice rather than corpus-speaker prompts. For each random split, all recordings of held-out test sentences are excluded from both TTS fine-tuning and Syl-BiLSTM training.

For objective translation quality, we follow the protocol in SeamlessExpressive \cite{seamless2023expressive} and compute the ASR-BLEU score by transcribing the generated audio using the Whisper Large model \cite{radford2023robust}. To evaluate speech naturalness and audio quality, we employ the UTMOS metric \cite{saeki2022utmos}. For cross-lingual stress preservation, we report our proposed Cross-lingual Emphasis Transfer Score (CETS) under both Word-Level (CETS-W) and Sentence-Level (CETS-S) matching conditions.

\begin{table*}[t]
    \centering
    \caption{Objective S2ST results. Metrics are Mean $\pm$ Std over 5 random test splits; bold indicates best performance.}
    \label{tab:s2st_objective}
    \begin{tabular}{l c c c c}
        \toprule
        \textbf{System} & \textbf{BLEU} ($\uparrow$) & \textbf{UTMOS} ($\uparrow$) & \textbf{CETS-W (\%)} ($\uparrow$) & \textbf{CETS-S (\%)} ($\uparrow$) \\
        \midrule
        Qwen2.5-Omni & 51.35 $\pm$ 3.08 & 3.53 $\pm$ 0.03 & 25.70 $\pm$ 6.00 & 21.10 $\pm$ 6.70 \\
        GPT-4o-audio & 50.63 $\pm$ 1.72 & 3.46 $\pm$ 0.04 & 25.20 $\pm$ 6.60 & 19.40 $\pm$ 7.10 \\
        Gemini + Base & \textbf{54.26 $\pm$ 2.91} & 3.60 $\pm$ 0.03 & 15.70 $\pm$ 4.60 & 16.60 $\pm$ 3.80 \\
        StressTransfer + Base & 46.34 $\pm$ 1.92 & 3.54 $\pm$ 0.05 & 24.60 $\pm$ 5.80 & 22.30 $\pm$ 5.50 \\
        \midrule
        \textbf{Proposed} & 47.35 $\pm$ 1.52 & \textbf{3.68 $\pm$ 0.02} & \textbf{60.80 $\pm$ 7.70} & \textbf{58.30 $\pm$ 6.90} \\
        \bottomrule
    \end{tabular}
\end{table*}

\subsection{Mandarin Stress Detection Evaluation}
To validate the efficacy of our proposed Syl-BiLSTM, we compare it against two baselines derived from EmphaClass \cite{de2024emphassess}, the state-of-the-art English emphasis detection model. 
Frame-Linear follows the core EmphaClass paradigm: final-layer XLS-R frame classification followed by majority voting over character boundaries. 
Frame-BiLSTM introduces a BiLSTM layer after the final XLS-R layer for contextual modeling, while maintaining frame-level classification with majority voting.

\begin{table}[H]
    \centering
    \caption{Performance comparison of stress detection models on Mandarin Chinese.}
    \label{tab:detection_results}
    \setlength{\tabcolsep}{4pt}
    \begin{tabular}{l c c c}
        \toprule
        \textbf{Architecture} & \textbf{Precision} & \textbf{Recall} & \textbf{F1-score} \\
        \midrule
        Frame-Linear & 0.78 $\pm$ 0.02 & 0.18 $\pm$ 0.04 & 0.29 $\pm$ 0.05 \\
        Frame-BiLSTM & 0.77 $\pm$ 0.02 & 0.58 $\pm$ 0.02 & 0.66 $\pm$ 0.01 \\
        \textbf{Syl-BiLSTM} & \textbf{0.94 $\pm$ 0.01} & \textbf{0.89 $\pm$ 0.03} & \textbf{0.91 $\pm$ 0.01} \\
        \bottomrule
    \end{tabular}
\end{table}

As shown in Table \ref{tab:detection_results}, frame-level EmphaClass-style baselines heavily degrade when applied to Mandarin. While Frame-BiLSTM improves over Frame-Linear by incorporating contextual dependencies, it still significantly lags behind our proposed method. Syl-BiLSTM achieves an F1-score of 0.91, demonstrating that our three key designs---syllable-level pooling, multi-layer feature fusion, and BiLSTM context encoding---are crucial for matching the linguistic and acoustic characteristics of Chinese stress.

\subsection{S2ST Objective Evaluation}
For the S2ST task, we benchmark our proposed system against five strong baselines:
1) Qwen2.5-Omni and 2) GPT-4o-audio-preview, representing the state-of-the-art open-source and closed-source end-to-end multimodal LLMs, respectively. Both were prompted to recognize source stress, translate, and synthesize the emphasized Chinese speech.
3) Gemini 2.5 Pro + CosyVoice3 Base, representing a highly capable cascaded system using Gemini for stress-annotated S2T and CosyVoice3 for TTS.
4) Base (StressTransfer + CosyVoice3 Base), which follows the architecture proposed in \cite{chen2025stresstransfer} without our TTS fine-tuning.
5) Proposed, our complete pipeline featuring the fine-tuned CosyVoice3 model.

To comprehensively assess stress preservation, we evaluate the performance using two distinct granularities of our metric:
\begin{itemize}
    \item \textbf{CETS-W (Word-Level Strict Transfer Rate):} Calculates the ratio of successfully transferred stressed words to the total number of stressed words in the source text. It evaluates the model's word-by-word emphasis accuracy.
    \item \textbf{CETS-S (Sentence-Level Full Transfer Rate):} A highly rigorous metric that considers an entire sentence successful if and only if \textit{all} of its source stressed words are accurately transferred. It evaluates the model's global prosodic consistency across the utterance.
\end{itemize}

The objective results are summarized in Table \ref{tab:s2st_objective}. Our proposed method shows stronger cross-lingual stress transfer than the baselines. Under the demanding CETS-S criterion, our model achieves 58.30\%, whereas the baseline systems remain around 16\%--22\%. This indicates that while baselines may occasionally generate isolated stress, they struggle to preserve sentence-level emphasis consistently.

Furthermore, while the billion-parameter LLMs (Gemini, Qwen) achieve slightly higher BLEU scores due to their strong text-translation capacities, our model maintains competitive translation quality compared with the direct baseline (StressTransfer + Base). Its highest UTMOS score also suggests that the TTS adaptation preserves audio naturalness while improving stress controllability.

\subsection{Subjective Evaluation and Metric Validation}
\label{subsec:subjective}
To further substantiate our findings, we conducted a subjective listening test. We randomly selected 20 samples from one test split and recruited 6 proficient bilingual speakers. The evaluators were asked to read the source English text (with stress highlighted), listen to the source audio, and then listen to the translated Chinese audio. They made a binary judgment (Success/Fail) based strictly on whether the emphasis was successfully transferred to the correct semantic position, explicitly ignoring overall translation and audio quality.

\begin{table}[H]
    \centering
    \caption{Subjective success rates for stress transfer.}
    \label{tab:subjective_results}
    \begin{tabular}{l c}
        \toprule
        \textbf{System} & \textbf{Success Rate (\%)} \\
        \midrule
        Qwen2.5-Omni & 11.67 $\pm$ 11.55 \\
        GPT-4o-audio & 25.00 $\pm$ 0.00 \\
        Gemini + Base & 16.67 $\pm$ 20.82 \\
        StressTransfer + Base & 16.67 $\pm$ 20.21 \\
        \midrule
        \textbf{Proposed} & \textbf{78.33 $\pm$ 2.89} \\
        \bottomrule
    \end{tabular}
\end{table}
\vspace{-0.25cm}

As shown in Table \ref{tab:subjective_results}, the subjective evaluation is consistent with the objective CETS metrics, with our proposed system leading by a large margin. To quantify this relationship, we performed a Pearson correlation analysis between the automatic pipeline evaluations and the subjective judgments (majority vote of 6 humans over 100 system outputs, i.e., 20 source samples across 5 systems). The analysis revealed a positive correlation ($r = 0.52, p < 10^{-6}$) and an absolute agreement rate of 79\%, suggesting that CETS is a useful automatic proxy for human perception in cross-lingual stress evaluation.

\section{Conclusion}
\label{sec:conclusion}

This paper introduced a framework for evaluating and preserving lexical stress in English-to-Chinese S2ST. By combining a Mandarin stress corpus, the Syl-BiLSTM detector, the CETS evaluation protocol, and stress-controllable TTS adaptation, our system improves cross-lingual emphasis transfer while maintaining translation quality and speech naturalness. Future work will extend the evaluation and generation framework to broader speakers and tonal-language settings.

\section{Acknowledgements}

This paper is supported by Project W2531054 of the National Natural Science Foundation of China, and the Program for Guangdong Introducing Innovative and Entrepreneurial Teams.

\section{Use of Generative AI Disclosure}

Generative AI tools were used for manuscript polishing. All scientific content, experimental results, and conclusions were reviewed and verified by the authors.

\bibliographystyle{IEEEtran}
\bibliography{mybib}

\end{document}